\documentclass[conference]{IEEEtran}
\IEEEoverridecommandlockouts
\usepackage{amsmath,amssymb}
\usepackage{newtxtext,newtxmath}
\usepackage{array}
\usepackage{balance}
\usepackage{booktabs}
\usepackage{cite}
\usepackage{graphicx}
\usepackage{microtype}
\usepackage{xspace}

\newcommand{\method}{CableMoE\xspace}
\newcommand{\vect}[1]{\boldsymbol{#1}}

\title{CableMoE: Harm-Regularized Mean-Residual Expert Fusion for\\
Transmission-Corridor Segmentation}

\author{\IEEEauthorblockN{Shima S. Salehi}
\IEEEauthorblockA{Department of Electrical and Computer Engineering\\
Texas A\&M University, College Station, TX, USA\\
shima.salehi@tamu.edu}}

\begin{document}
\maketitle

\IEEEpubid{\makebox[\columnwidth]{979-8-3315-5720-1/26/\$31.00~\copyright~2027 IEEE\hfill}%
\hspace{\columnsep}\makebox[\columnwidth]{}}
\enlargethispage{-2\baselineskip}

\begin{abstract}
Semantic mapping of transmission corridors benefits from heterogeneous LiDAR
and image-guided predictors, but replacing a strong ensemble with a learned
router can damage rare electrical assets. We introduce \method, a post-hoc
mixture of four frozen segmentation experts. Its deployable prediction is a
calibrated residual around availability-aware probability averaging, while a
harm penalty discourages extra loss on points that the mean already classifies
correctly. A matched component study shows why each design choice is needed:
an unconstrained learned mixture loses 1.032 mIoU points relative to probability
mean, harm regularization restores 0.359 points at fixed capacity, and removing
an unnecessary point-wise gate slightly improves accuracy and stability. With
three zones for fitting, two disjoint zones for residual-strength selection,
and router-untouched t5b5 for final reporting, \method obtains
$84.906\!\pm\!0.099$\% mIoU and $98.140\!\pm\!0.008$\% overall accuracy. This is
$+10.000$ points over SPT and $+0.263$ over probability mean; all three seeds
improve the latter. Across pylon, conductor, structural-cable, and insulator
classes, macro-IoU rises from 84.517\% for SPT and 97.153\% for probability mean
to $98.144\!\pm\!0.062$\%. The results support conservative residual composition,
not a claim of a new segmentation backbone or universal state of the art.
\end{abstract}

\begin{IEEEkeywords}
power-line inspection, point-cloud segmentation, multimodal fusion,
mixture of experts, LiDAR
\end{IEEEkeywords}

\section{Introduction}

Reliable inspection planning begins with a reliable inventory of the corridor.
Airborne mapping can provide this inventory at scale, but its semantic map must
retain thin, sparsely sampled conductors, structural cables, pylons, and
insulators while separating them from dominant vegetation and terrain. LiDAR
captures geometry reliably, whereas registered imagery can resolve appearance
ambiguities; imagery, however, may be occluded or unavailable outside the
camera footprint. The resulting problem is therefore both class-imbalanced and
sensor-dependent.

\begin{figure}[t]
\centering
\includegraphics[width=\columnwidth]{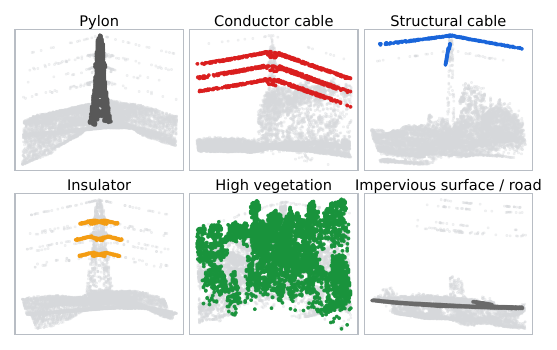}
\caption{Representative GridNet-HD point-cloud crops from the grouped
taxonomy. Color marks the named target class and gray gives neighboring scene
context. These are labeled dataset examples, not model predictions.}
\label{fig:dataset_samples}
\end{figure}

AI applications in power systems span temporal monitoring and spatial mapping.
For example, GAN--LSTM models detect anomalous smart-meter events over time
\cite{nia2026ganlstm}; here, the objective is instead a dense spatial inventory
in which every LiDAR point receives an infrastructure or context label.
GridNet-HD makes this setting measurable by providing aligned high-density
LiDAR, imagery, eleven grouped classes, and heterogeneous frozen predictions
\cite{carreaud2026gridnethd}. Its asset and context classes differ sharply in
geometry and support, as illustrated by the labeled crops in
Fig.~\ref{fig:dataset_samples}.

The available experts reflect successive advances in 3-D understanding, from
hierarchical point sets~\cite{qi2017pointnetpp}, deformable point convolutions
\cite{thomas2019kpconv}, and sparse voxel--point processing
\cite{tang2020spvcnn} to SPT's hierarchical superpoints
\cite{robert2023spt} and PTv3's serialized point attention
\cite{wu2024ptv3}. Image voting and late fusion add complementary appearance
evidence. Yet no expert dominates every class or location: their failures are
complementary, their confidence scales are not directly comparable, and errors
on rare electrical assets can disappear inside overall accuracy.

This complementarity changes the research question. The immediate need is not
another segmentation backbone, but a dependable way to compose already-trained
experts. A simple probability mean is a surprisingly strong control---in our
geographic evaluation it exceeds SPT by 9.737 mIoU points. Directly replacing
that mean with a learned router can therefore destroy useful consensus while
trying to exploit expert specialization. A learned composer should instead
justify and limit every departure from the strong average.

We address this requirement with \method, a deliberately post-hoc
\emph{mean-residual} mixture of four frozen experts. Availability-aware
probability mean remains inside the deployment rule; a small router proposes
only a convex direction of change. A scalar selected exclusively on calibration
zones limits the residual strength, while a harm term penalizes changes that
corrupt points already classified correctly by the mean. Freezing every expert
isolates the value of composition from backbone retraining and permits an exact,
source-order-preserving implementation.

Our contributions are:
\begin{itemize}
  \setlength{\itemsep}{0pt}
  \item a probability-mean-contained convex MoE that combines availability
  masking with an explicit penalty for harming a strong ensemble baseline;
  \item an exact-identity implementation that preserves source-LAS order while
  composing frozen LiDAR and image-guided predictions;
  \item a disjoint three-zone fit, two-zone calibration, and router-untouched
  geographic evaluation with three seeds; and
  \item a matched component study that separates probability containment and
  harm regularization, rejects an unsupported point-wise gate, and reports
  asset-specific behavior.
\end{itemize}

\begin{figure*}[!t]
\centering
\includegraphics[width=\textwidth]{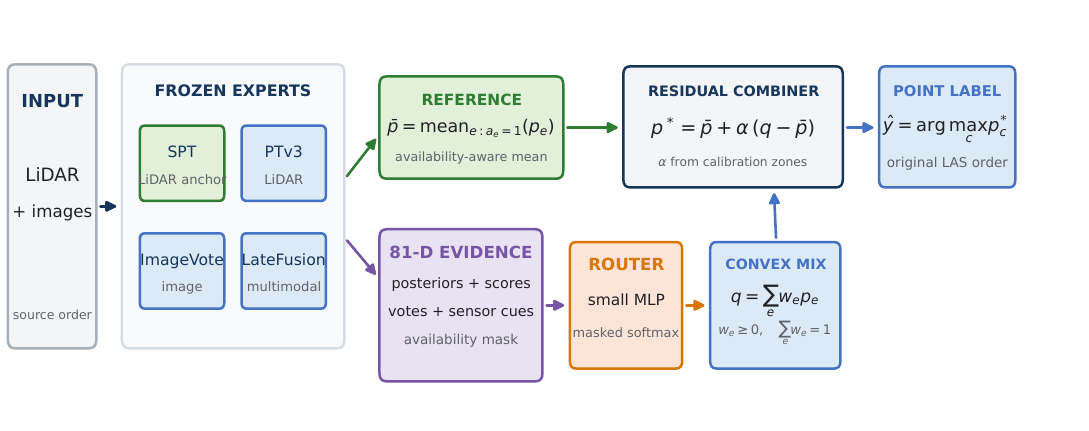}
\caption{Final \method pipeline. Four frozen experts preserve source-point
identity. Availability-aware probability mean is the contained baseline. An
81-D evidence vector drives a small MLP that predicts convex expert weights.
The final distribution adds a calibration-selected, harm-regularized residual;
there is no learned point-wise gate.}
\label{fig:pipeline}
\end{figure*}

\section{Related Work}

\textbf{Transmission-corridor segmentation.}
GridNet-HD supplies aligned airborne LiDAR, oblique imagery, eleven grouped
classes, and frozen baseline outputs~\cite{carreaud2026gridnethd}. TS40K,
ECLAIR, and TowerDataset provide related rural, aerial, and long-corridor
benchmarks~\cite{lavado2025ts40k,melekhova2024eclair,cui2026towerdataset}, but
their taxonomies and protocols differ. Recent corridor methods emphasize
critical regions~\cite{pei2025tlsnet}, point attention
\cite{lu2025emaflptv3,wu2026tlpt}, graph context~\cite{zhang2025gkcae},
elevation~\cite{wang2025empowernet}, geometry~\cite{lavado2026gibly}, and class
imbalance~\cite{zhang2026pointgrid}. DITR transfers 2-D foundation features
into 3-D~\cite{knaebel2026ditr}, while DeepChoice learns image-view
weighting~\cite{carreaud2026deepchoice}. More general multimodal segmentation
either performs bidirectional 2-D/3-D projection~\cite{hu2021bpnet} or distills
image priors into a LiDAR-only model~\cite{yan2022twodpass}. Local neighborhood pooling provides a
complementary way to aggregate texture and context in remote-sensing imagery
\cite{nia2026nfp}, while traffic-aware intention prediction demonstrates
context-conditioned graph reasoning in a different spatiotemporal setting
\cite{nia2025traffic}. These methods train representations or task-specific
predictors; we instead compose already-trained heterogeneous segmenters.

Thin electrical assets also connect this problem to category-specific remote
sensing fusion~\cite{zhu2026classwisecrf}, spectral point representations
\cite{wen2023pointwavelet}, topology-preserving objectives for tubular
structures~\cite{shit2021cldice}, and dedicated airborne-laser power-line
extraction~\cite{cao2025powerline}. An end-to-end phenotyping pipeline further
illustrates how image extraction, analysis, and evaluation can be designed as
one reproducible scientific workflow~\cite{nia2026phenotyping}. These ideas are
complementary to our study:
they can improve an individual expert or impose a structural prior, whereas
our controlled question is whether frozen, heterogeneous experts can be
composed without sacrificing a strong probability-average baseline.

\textbf{Expert composition and conservative prediction.}
Stacked generalization and dynamic classifier selection combine diverse
predictors~\cite{wolpert1992stacked,woods1997dynamic}. Mixture-of-experts and
learning-to-defer methods further learn which predictor to trust
\cite{shazeer2017moe,mozannar2020defer,verma2023multiple,mao2023twostage}.
Deep ensembles show that averaging independently trained predictors can
improve accuracy and uncertainty quality~\cite{lakshminarayanan2017deep}; our
probability mean differs only in that its members are heterogeneous frozen
architectures. Selective prediction
often trades accuracy against coverage~\cite{geifman2019selectivenet}, whereas
our output is always a dense semantic label. Calibration remains difficult for
imbalanced segmentation~\cite{wang2023calibration}; therefore, we do not treat
raw confidence as a risk guarantee. Our narrower contribution is baseline
containment: learning only a calibrated residual around probability mean and
penalizing loss increases on mean-correct points.

Related 3-D reconstruction studies use local/global wavelet regularization to
stabilize sparse-view Gaussian splatting~\cite{salehi2026lgdwtgs} and
motion-regime experts to specialize dynamic Gaussian predictors
\cite{salehi2026moregs}. Those methods optimize scene representations or
motion-specific predictors; in contrast, \method performs post-hoc composition
of frozen semantic segmenters while retaining a strong ensemble baseline.

\section{Method}

\subsection{Frozen experts and contained baseline}

For source point $i$, expert $e\in\{1,\ldots,4\}$ provides an 11-class
distribution $\vect p_{i,e}$. The experts are SPT, PTv3, ImageVote, and
LateFusion. An availability indicator $a_{i,e}$ masks missing image views and
missing sidecar predictions. The strongest no-training composition in our
geographic test is
\begin{equation}
\bar{\vect p}_i=\frac{\sum_e a_{i,e}\vect p_{i,e}}
{\sum_e a_{i,e}}.
\label{eq:mean}
\end{equation}
Every array is indexed by zero-based source-LAS ordinal. We do not use
coordinate matching or nearest-neighbor reassignment, eliminating a common
source of silent label transport errors.

\subsection{Evidence and convex expert mixture}

The 81-D evidence concatenates 44 class probabilities, four availability
flags, per-expert confidence, margin, and entropy (12), an 11-bin vote
histogram, and ten sensor/context values: height, intensity, RGB, returns, scan
angle, and local 25-m-block position. Absolute coordinates and zone identity
are excluded. A LayerNorm--GELU MLP
$81\!\rightarrow\!160\!\rightarrow\!96\!\rightarrow\!4$ predicts expert
logits. Masked softmax yields $w_{i,e}\ge0$ and
$\sum_e w_{i,e}=1$, giving
\begin{equation}
\vect q_i=\sum_e w_{i,e}\vect p_{i,e}.
\label{eq:mix}
\end{equation}

\subsection{Harm-regularized mean residual}

The final distribution is
\begin{equation}
\vect p_i^*=\bar{\vect p}_i+
\alpha(\vect q_i-\bar{\vect p}_i),\quad \alpha\in[0,1].
\label{eq:residual}
\end{equation}
Because both terms are distributions, Eq.~\eqref{eq:residual} is convex. At
$\alpha=0$, it is exactly probability mean; hence the learned router cannot
erase the baseline from the policy definition. For each seed, calibration
chooses $\alpha\in\{0.25,0.5,0.75,1\}$ before evaluation labels are read.

Let $\ell_i=-\log q_i(y_i)$ and
$\bar\ell_i=-\log\bar p_i(y_i)$. Training minimizes
\begin{equation}
\mathcal L_i=\ell_i+
2\mathbb 1[\arg\max\bar{\vect p}_i=y_i]
[\ell_i-\bar\ell_i]_++0.1H(\bar{\vect p}_i,\vect q_i).
\label{eq:loss}
\end{equation}
The second term penalizes extra loss precisely when the strong mean is already
correct; the last term distills its full distribution. Class-balanced sample
weights multiply Eq.~\eqref{eq:loss}. The final model has no point-wise
residual gate: the ablation in Section~\ref{sec:results} shows that the gate is
unnecessary, and its removal reduces both complexity and seed variance.

\subsection{Matched component variants}

We test four trainable variants under identical zones, features, seeds, and
optimization. \emph{Pure learned MoE} uses $\vect q$ directly and removes both
harm and distillation. \emph{No harm} retains the learned gate and distillation
but sets the harm weight to zero. \emph{Harm + gate} is the earlier full
parameterization. \emph{Final harm, no gate} uses Eqs.~\eqref{eq:residual}--
\eqref{eq:loss}. Probability mean is a deterministic, no-training control.
Thus the comparisons isolate whether mean containment, harm regularization,
and gating are supported by held-out geography.

\section{Experimental Protocol}

\subsection{Data roles and training}

GridNet-HD has eleven grouped classes. The released SPT, PTv3, ImageVote, and
LateFusion probabilities are stored in float16 and converted to normalized
float32 distributions before composition. The six labeled public zones belong
to the benchmark's official validation release and are excluded from expert
training, although aggregate validation performance may have influenced
released checkpoint selection. We therefore claim router-untouched evaluation,
not a completely unseen expert domain.

The geographic protocol assigns t1z5b, t4z1, and t5b1 to router fitting;
t1z8 and t3z4 to hyperparameter calibration; and t5b5 to final evaluation.
Deterministic source-ordinal hashing retains 423,231 disagreement-focused fit
samples and 396,178 calibration candidates. No t5b5 label selects a weight,
seed, or residual strength. The evaluation contains 61,412,436 points and ten
present classes; water is absent and is omitted rather than assigned zero IoU.

All variants use AdamW for 15 epochs, batch size 32,768, learning rate
$6\times10^{-4}$, weight decay $3\times10^{-4}$, and gradient norm 5. Three
seeds (20261105--20261107) share identical data roles. Final deployment uses
$\alpha=0.75$, selected on the two calibration zones. The router is small, but
end-to-end latency includes four expert inferences and is not measured.

\subsection{Metrics and claim boundary}

Primary metric is ground-truth-present mIoU; overall accuracy (OA) and
per-class IoU are secondary. Mean and population standard deviation summarize
three router fits on the same evaluation population; they measure fitting
stability, not geographic confidence. Comparisons to methods trained with
different data or preprocessing are contextual. We make no state-of-the-art,
formal risk-control, fault-detection, clearance, or broad-domain claim.

\section{Results}
\label{sec:results}

\subsection{Geographic evaluation}

Table~\ref{tab:deployment} reports the frozen experts, strongest simple
composition, and final model on router-untouched t5b5. Probability mean already
reaches 84.643\% mIoU, 9.737 points above SPT. \method reaches
$84.906\pm0.099$\%, a gain of $10.000\pm0.099$ over SPT and
$0.263\pm0.099$ over probability mean. The three seed results are 85.040,
84.875, and 84.803\%; every fit exceeds the deterministic mean. OA also rises
from 98.039\% to $98.140\pm0.008$\%. These are deliberately modest claims over
the strongest control: the large improvement over SPT is primarily due to
expert complementarity, while the learned residual supplies the final
increment.

\begin{table}[t]
\centering
\caption{Router-untouched t5b5 evaluation. Mean $\pm$ population SD uses three
fits; $\alpha$ is selected on t1z8/t3z4 only.}
\label{tab:deployment}
\small
\setlength{\tabcolsep}{4pt}
\begin{tabular}{lrr}
\toprule
Method & mIoU (\%) & OA (\%)\\
\midrule
SPT anchor & 74.906 & 97.407\\
PTv3 & 74.739 & 97.575\\
LateFusion & 79.081 & 92.964\\
Probability mean & 84.643 & 98.039\\
\textbf{Final \method} & $\mathbf{84.906\pm0.099}$ & $\mathbf{98.140\pm0.008}$\\
\bottomrule
\end{tabular}
\end{table}

\subsection{What each component contributes}

\begin{figure*}[!t]
\centering
\includegraphics[width=0.98\textwidth]{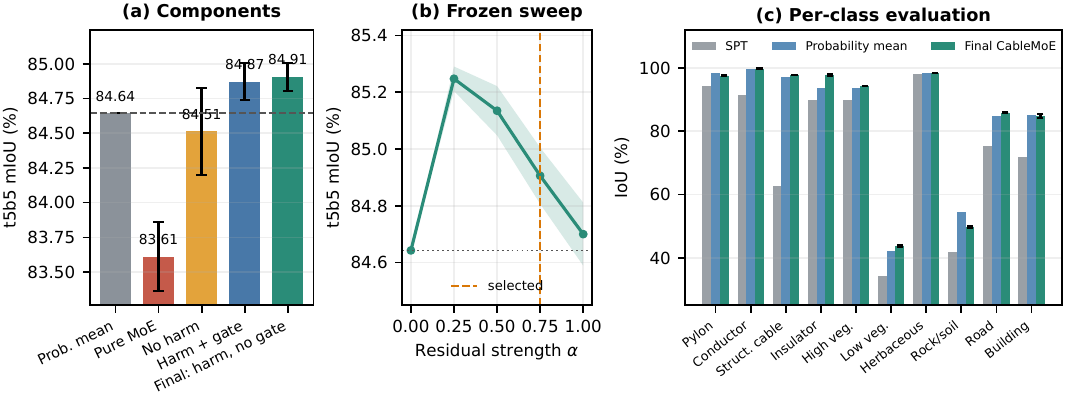}
\caption{Final evaluation. (a) Matched component ablation. (b) Frozen
residual-strength sweep; the dashed line marks the calibration-selected
$\alpha=0.75$ and the dotted line is probability mean. (c) Per-class IoU for
SPT, probability mean, and final \method. Error bands/bars show one population
SD over three fits. Water is absent from t5b5.}
\label{fig:ablation}
\end{figure*}

Figure~\ref{fig:ablation}(a) answers the central
design question. Replacing the mean with a pure learned MoE reduces mIoU by
1.032 points, so probability averaging is not an expendable baseline. With the
same learned gate and distillation, adding the harm term improves 84.513 to
84.872 mIoU ($+0.359$). Removing the gate then slightly improves the mean by
0.034 and reduces SD from 0.135 to 0.099. The gate's contribution is therefore
unsupported; the final architecture omits it. This matched evidence is also
why we do not present an increasingly complex router as the novelty.

\subsection{Residual-strength sensitivity}

Figure~\ref{fig:ablation}(b) evaluates the complete predeclared $\alpha$ grid
after the calibration choice is frozen. Every nonzero candidate remains above
probability mean on t5b5. The evaluation means are 85.247, 85.135, 84.906, and
84.700\% for $\alpha=0.25,0.5,0.75,1$, respectively. Thus calibration-selected
$\alpha=0.75$ is robust but is not the retrospective t5b5 optimum. We keep the
pre-evaluation choice and do not report 85.247\% as the method's final score.
This gap is evidence that residual strength does not transfer perfectly across
geographies and should be selected with more calibration zones in future work.

\subsection{Electrical assets and failure modes}

Figure~\ref{fig:ablation}(c) separates the electrical classes that matter for
corridor inventory. Relative to SPT, final \method improves pylon, conductor,
structural-cable, and insulator IoU by 3.334, 8.259, 35.015, and 7.903 points.
Their macro-IoU increases from 84.517\% (SPT) and 97.153\% (probability mean)
to $98.144\pm0.062$\%. Relative to the already strong mean, the residual adds
0.027 on conductor, 0.701 on structural cable, and 3.962 on insulator, but
reduces pylon by 0.725. It also improves low vegetation by 1.780 and road by
1.165, while rock/soil and building fall by 4.628 and 0.475. Thus the small
aggregate gain is not uniformly distributed. The observed rock/soil trade-off
is the main remaining failure mode and motivates class-aware residual control,
but tuning it after t5b5 would invalidate the present protocol.

\begin{figure*}[!t]
\centering
\includegraphics[width=0.96\textwidth]{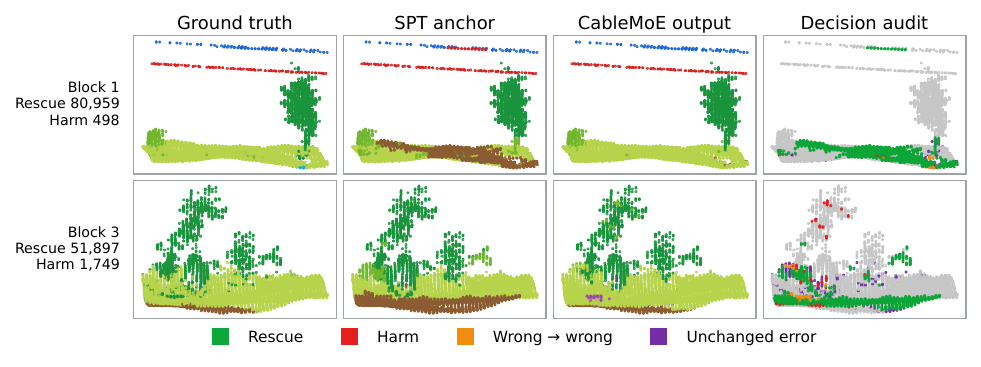}
\caption{Qualitative mechanism audit on two held-out spatial development
blocks. Columns show ground truth, SPT, the routed development output, and
point-level edit outcomes. Green denotes rescue, red harm, orange a changed
but still-wrong label, and purple an unchanged error. These examples visualize
the edit mechanism only; they are not the source of the final t5b5 numbers in
Table~\ref{tab:deployment}.}
\label{fig:qualitative}
\end{figure*}

\subsection{Frozen expert-outage stress test}

Table~\ref{tab:outage} tests the frozen model without outage-specific fitting
or calibration. We retain the same 61,412,436-point evaluation population and
disable expert posteriors only through their availability masks. The full row
exactly reproduces Table~\ref{tab:deployment}. With ImageVote absent, \method
remains 0.556 mIoU points above the matched probability mean. This behavior
does not extend to every outage: removing LateFusion yields a 0.244-point
deficit, and retaining only the two LiDAR experts yields a 1.193-point deficit.
Availability masking therefore handles missing inputs mechanically but does
not guarantee robustness to an unseen availability distribution.

\begin{table}[t]
\centering
\caption{Post-freeze expert-outage stress test on t5b5. Mean $\pm$ population
SD uses three fits; $\Delta$ is relative to the availability-matched mean.}
\label{tab:outage}
\small
\setlength{\tabcolsep}{2.6pt}
\begin{tabular}{lrrr}
\toprule
Available setting & Mean & \method & $\Delta$\\
\midrule
Full bank & 84.643 & $84.906\pm0.099$ & $+0.263$\\
No ImageVote & 83.125 & $83.681\pm0.208$ & $+0.556$\\
No LateFusion & 83.298 & $83.054\pm0.161$ & $-0.244$\\
LiDAR-only & 79.921 & $78.727\pm0.383$ & $-1.193$\\
\bottomrule
\end{tabular}
\end{table}

On an eight-thread x86 CPU, LAS reading and shared preprocessing, evidence
construction, and the no-gate mixer require 22.1, 68.3, and 38.0 seconds,
respectively, for 63.5 million raw points. The 128.4-second composition path
corresponds to 2.02~$\mu$s/point (0.495 million points/s). This is post-hoc
fusion overhead; it excludes the substantially larger and hardware-dependent
cost of running the four frozen experts.

\section{Discussion}

\textbf{Why the result is useful.}
The experiment shows that most performance comes from heterogeneous expert
composition, not a deeper router. The final contribution is a conservative way
to learn on top of that strong composition: retain probability mean in the
policy, penalize mean-correct damage, and select only one residual-strength
scalar outside the evaluation zone. This is directly relevant to corridor
mapping because the four electrical-asset classes improve strongly over the
single LiDAR anchor. The output remains a semantic inventory layer; geometric
clearance and condition assessment require separately validated geometry and
topology.

\textbf{Why a residual rather than direct replacement.}
A conventional gate is free to replace the ensemble distribution everywhere,
including regions where averaging already suppresses an expert-specific
mistake. Equation~\eqref{eq:residual} instead separates two roles: the mean is
the always-present reference and the router estimates only a direction of
change. The scalar $\alpha$ then limits how far deployment can move along that
direction. This is not a formal safety bound, but it makes the intervention
strength observable and testable; Fig.~\ref{fig:ablation}(a) empirically confirms
that unconstrained replacement is the weakest matched trainable variant.

\textbf{Limitations.}
All six public zones come from one benchmark release, and three seeds do not
measure geographic uncertainty. t5b5 is untouched by the router protocol but
is not an expert-level external dataset. The gain over probability mean is
small, and three context classes regress. No matched official hidden score is
available for this final no-gate model; an older hard-selector leaderboard
result is intentionally excluded because it evaluates a different policy.
Four frozen experts are also more expensive than a single backbone. The
post-hoc mixer is inexpensive, but end-to-end expert latency, energy, and
sensor shift remain unmeasured. The outage stress test shows that availability
masking alone is insufficient for LateFusion loss and LiDAR-only operation;
expert-dropout training requires evaluation on a new untouched zone.
The harm loss is an empirical regularizer, not a conformal or probabilistic
risk guarantee.

\textbf{Reproducibility.}
The final expert order is SPT/PTv3/ImageVote/LateFusion. Zone roles, source
ordinals, availability masks, class mapping, three seeds, and calibration grid
are fixed before evaluation. The calibration-selected deployment checkpoint is
seed 20261107 with $\alpha=0.75$; the three-seed table, rather than this single
checkpoint, supports the reported stability. All metrics are recomputed from
raw-order predictions, and absent classes are recorded explicitly.

\section{Conclusion}

\method composes four frozen transmission-corridor segmenters using a
harm-regularized convex residual around probability mean. A matched ablation
shows that baseline containment and the harm term are useful, while a learned
point-wise gate is not. On router-untouched t5b5, the final no-gate model obtains
$84.906\pm0.099$\% mIoU and $98.140\pm0.008$\% OA, improving SPT by 10.000
points and the strong probability mean by 0.263. The method is therefore best
viewed as a conservative expert-composition pipeline, not a new backbone or a
universal risk-control mechanism. A matched hidden evaluation and additional
geographies are the highest-value next tests.

\section*{Acknowledgment}

{ Texas A\&M HPRC provided computing resources. AI assisted with manuscript-wide language.}

\bibliographystyle{IEEEtran}
\bibliography{bibliography/references,bibliography/cablemoe_refs}

\end{document}